\documentclass[journal,12pt,onecolumn,draftclsnofoot]{IEEEtran}
\usepackage{eurosym,bm}
\usepackage{cite,algorithm,array,url}
\usepackage{graphicx,colortbl}
\usepackage{amsfonts,times,amsmath,amssymb,amsthm}
\usepackage{multirow,booktabs}
\usepackage{color}

\ifCLASSOPTIONcompsoc
\usepackage[caption=false,font=normalsize,labelfon
t=sf,textfont=sf]{subfig}
\else
\usepackage[caption=false,font=footnotesize]{subfi
g}
\fi

\graphicspath{{images/}}

\theoremstyle{plain}

\theoremstyle{definition}

\newtheorem{assumption}{Assumption}

\theoremstyle{remark}
\newtheorem{remark}{Remark}

\begin{document}

\title{A Learning-Based Computational Impact Time Guidance}
\author{Zichao~Liu, Jiang~Wang, Shaoming~He\textsuperscript{*}, Hyo-Sang~Shin and Antonios~Tsourdos\thanks{
Zichao Liu, Jiang Wang and Shaoming He are with the School of Aerospace Engineering, Beijing Institute of Technology, Beijing 100081, China}
\thanks{
Hyo-Sang Shin and Antonios Tsourdos are with the School of Aerospace, Transport and Manufacturing, Cranfield University, Cranfield MK43 0AL, UK}
\thanks{\textsuperscript{*}Corresponding Author. Email: $\texttt{shaoming.he@bit.edu.cn}$}}
\maketitle

\begin{abstract}
This paper investigates the problem of impact-time-control and proposes a learning-based computational guidance algorithm to solve this problem. The proposed guidance algorithm is developed based on a general prediction-correction concept: the exact time-to-go under proportional navigation guidance with realistic aerodynamic characteristics is estimated by a deep neural network and a biased command to nullify the impact time error is developed by utilizing the emerging reinforcement learning techniques. The deep neural network is augmented into the reinforcement learning block to resolve the issue of sparse reward that has been observed in typical reinforcement learning formulation. Extensive numerical simulations are conducted to support the proposed algorithm.
\end{abstract}



\section{Introduction}

The primary objective of missile guidance law is to guide the vehicle to intercept the target with zero miss distance \cite{lee2019capturability}. The most widely-used proportional navigation guidance (PNG) law enjoys the merit of easy implementation and provides the possibility of energy minimization under certain circumstances \cite{jeon2010optimality,cho2016optimality}. The PNG has also been proved to maximize the terminal velocity in a recent work \cite{jeon2019connections}. However, conventional PNG cannot handle additional constraints, e.g., impact angle, impact time, and needs further adjustments. Among these constraints, the impact-time-control guidance attracts extensive interests in recent years since this strategy helps to improve the probability of successful penetration against close-in weapon systems equipped on land or sea platforms. Generally, impact time control can be achieved by coordinating the predicted time-to-go through communication among the missile network or control the predicted time-to-go for each missile individually by a proper guidance law.

The impact-time-control guidance (ITCG) law was first introduced in \cite{jeon2006impact} by standard optimal control theory. Based on the concept of biased PNG, the authors  in \cite{kim2013biased} developed a polynomial biased term to cater for the impact time constraint. A generalized method that can be leveraged in extending existing guidance laws to impact time control was proposed in \cite{tahk2018impact} by utilizing a specific error dynamics. This concept was further extended to a three-dimensional engagement scenario in \cite{he2019three}. The work in \cite{cho2016modified} derived the analytic time-to-go estimation without any small angle assumption and hence can improve the performance of biased PNG for impact time control with large heading error scenarios. Except for PNG and its variants, geometric rules \cite{tsalik2019circular,zadka2020consensus,wang2019new} and nonlinear control theories \cite{sinha2020threea,chen2019sliding,kumar2015impact,sinha2020three,lee2020impact,saleem2016lyapunov,harl2011impact}, are also utilized in impact time control or coordination in recent works. The main idea behind these guidance algorithms is to adjust the remaining flight time, i.e., length of the trajectory, to control the intercept time. However, exact time-to-go estimation is intractable for real implementations and hence these algorithms leveraged approximate estimations by using small angle assumptions. Although guidance laws without explicit time-to-go estimation were also proposed in the literature \cite{he2017consensus,kim2018backstepping,kim2018sliding,tekin2016control,tekin2017polynomial}, most of them require constant-speed assumption in guidance command derivation. This means that the performance in impact time control will degrade in practical scenarios. Therefore, classical closed-form guidance laws that rely on approximated models with small angle or other idealistic assumptions, are no longer appealing to solve future real-world guidance problems. 

Thanks to the rapid development on embedded computational capability, there has been an increasing attention on the development of computational guidance algorithms in recent years \cite{lu2017introducing,kang2018optimal,liu2016closed,guo2020data}. Unlike classical optimal guidance laws, computational guidance algorithms generate the guidance command relies extensively on onboard computation and therefore dose not require analytic solution of specific guidance laws. Generally, computational guidance can be classified into two main categories: (1) model-based ; and (2) data-based. One of the most widely-used model-based computational missile guidance algorithms, termed as model predictive static programming (MPSP) \cite{dwivedi2011suboptimal,oza2012impact,hong2019model,hong2019smooth}, converts a dynamic programming problem into a static programming problem, thereby providing appealing characteristics in terms of computational efficiency \cite{padhi2009model}. However, the major limitation of MPSP-based computational guidance algorithms is that they require a good initial solution guess to guarantee the convergence \cite{pan2019newton,ma2020parallel}. 

Notice that impact time control requires accurate estimation of the remaining flight time. This requires finding the relationship between the predicted impact time and the nonlinear dynamic model. Hence, model-based impact-time-control guidance algorithms inevitably require computationally-expensive numerical integration in implementation. Another bottleneck of developing model-based impact-time-control algorithms is that the analytic dynamics model of the time-to-go (i.e., in terms of the lateral acceleration) is unknown. For this reason, the data-based model-free concept is more suitable for impact time control. Motivated by this observation, this paper aims to propose a computational impact-time-control guidance algorithm by leveraging the emerging deep learning techniques. The proposed guidance algorithm is developed based on a general prediction-correction concept: the exact time-to-go under PNG considering aerodynamic forces is estimated by a deep neural network (DNN) based supervised learning functional block; and a biased command for impact time control is developed by utilizing the state-of-the-art reinforcement learning (RL) approaches. Extensive numerical simulations with comparisons are also carried out to verify the effectiveness of the proposed approach.

The main contributions of this paper are twofolds. On one hand, we propose a general prediction-correction-based framework for computational guidance design. This concept can be easily extended to other guidance application scenarios, e.g., impact angle control, terminal velocity control. Up to the best of our knowledge, no similar results have been published in the existing literature. On the other hand, the supervised learning technique is augmented into the RL block to resolve the issue of sparse reward in RL training. This concept is demonstrated to significantly improve the learning efficiency.

The remainder of this paper is organized as follows. The backgrounds and preliminaries of this paper are stated in Sec. \ref{sec:1}. Section \ref{sec:2} presents the concept of the proposed computational guidance algorithm. Sec. \ref{sec:3} provides prediction of time-to-go, followed by the correction of impact time error in Sec. \ref{sec:4}. Finally, some simulation results and conclusions are offered. The source code of this paper is available at \texttt{https://github.com/LutterWa/Computational-ITCG-with-PPO}.

\section{Backgrounds and Preliminaries}
\label{sec:1}

In this section, we first present the dynamics and kinematics models of the interceptor. Then, the problem formulation of this paper is stated. Before presenting the mathematical models, we make the following general assumptions that have been widely-accepted in impact-time-control guidance law design.
\begin{assumption}
Since the control loop is generally much faster than that of the guidance loop, we assume that the missile's autopilot is ideal, i.e., there is no control delay.
\end{assumption}
\begin{assumption}
The target is assumed to be stationary due to the fact the concept of simultaneous attack is generally leveraged to intercept high-value ships or ground-based targets.
\end{assumption}

\begin{figure}[hbt!]
	\centering
	\includegraphics[width=.5\textwidth]{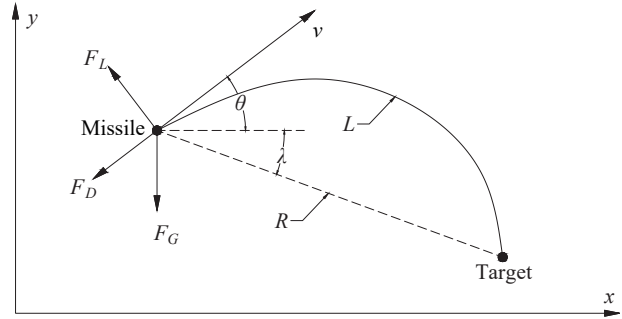}
	\caption{Definition of notations and symbols.}
	\label{fig:1}
\end{figure}

\subsection{Nonlinear Mathematical Models}

To simplify the problem, this paper only considers the vertical engagement geometry, as shown in Fig. \ref{fig:1}. The symbol $v$ stands for the missile's velocity and $ \theta $ represents the flight path angle. The lift, drag, and gravity forces are denoted by $ F_L $, $ F_D $, and $ F_G $. The notation $ \lambda $ is the line-of-sight (LOS) angle, and $ L $ stands for the length of the flight trajectory. The relative distance is denoted by $R$. The aerodynamic forces can be expressed as follows
\begin{gather}
	\label{equation:1}
	F_L=C_{L}QS\\
	\label{equation:2}
	F_D=C_{D}QS\\
	\label{equation:3}
	F_G=mg
\end{gather}
where $C_L$ denotes the lift coefficient and $ C_D $ stands for the drag coefficient. The symbol $S$ is the reference area of the missile and $m$ represents the mass of the missile. The notations $g$ stands for the gravitational acceleration and the variable $Q$ denotes the dynamic pressure, which can be determined by
\begin{equation}
	Q=\frac{1}{2}{\rho}v^{2}
\end{equation}
where $\rho$ stands for the air density.

The differential equations of the dynamics and kinematics models are given by
\begin{gather}
	\label{equation:5}
	\dot{v} =\frac{F_D - F_G\sin \theta}{m} \\
	\label{equation:6}
	\dot{\theta } =\frac{F_L - F_G\cos \theta}{mv} \\
	\label{equation:7}
	\dot{x} =v\cos \theta \\
	\label{equation:8}
	\dot{y} =v\sin \theta
\end{gather}
where $\left(x,y\right)$ denotes the inertial position of the missile.

Since the angle-of-attack (AoA), denoted by $ \alpha $, is a small variable, the aerodynamic coefficients can be approximated as
\begin{gather}
	\label{equation:9}
	C_{L} =C^{\alpha }_{L} \alpha \\
	\label{equation:10}
	C_{D} =C_{D0}+C^{\alpha ^{2}}_{D0} \alpha ^{2}
\end{gather}
where $ C^{\alpha }_{L} $ denotes the derivative of lift coefficient with respect to AoA and $ C_{D0} $ represents the parasite drag coefficient. The notation $ C^{\alpha ^{2}}_{D0} $ stands for the induced drag coefficient. These aerodynamic coefficients can be obtained through ground wind tunnel experiment. In evaluating the lift and drag forces, we leverage the standard atmosphere model to calculate the air density.

Since guidance laws normally generate lateral acceleration command $a_M$, we require the auxiliary relationship between AoA and lateral acceleration, i.e.,
\begin{equation}
	\alpha=\frac{m a_M}{C^{\alpha }_{L}  QS}
\end{equation}

\subsection{Impact Time Control Problem}

The objective of impact time control is to guide the missile to intercept the target at a prescribed time. Define $t_d$ as the desired interception time, the impact time constraint can then be formulated as
\begin{equation}
	\label{equation:11}
	t_{f}=t_d
\end{equation}
where $ t_f $ denotes the terminal time.

Since
\begin{equation}
t_f = t_{go} + t
\end{equation}
the problem of impact time control can be converted into an equivalent problem of adjusting the remaining flight time. Mathematically, the time-to-go can be readily formulated as
\begin{equation}
	\label{equation:13}
	t_{go} =\int ^{t_{f}}_{t}\frac{L}{v( \tau)} d\tau
\end{equation}
where the length of the trajectory is determined as
\begin{equation}
	\label{equation:14}
	L=\int_{x}^{x_{f}} \sqrt{1+\theta^{2}(x)} \mathrm{d} x
\end{equation}

A perfect interception requires
\begin{equation}
	\label{equation:12b}
{x} \left( t_f \right) = x_T, \quad {y} \left( t_f \right) = y_T
\end{equation}
where $\left(x_T,y_T\right)$ denotes the inertial position of the stationary target.

In summary, the main objective if this paper is to propose a computational guidance algorithm to satisfy constraints \eqref{equation:11} and \eqref{equation:12b}.

\begin{remark}
Notice that the aerodynamic forces are time-varying and hence finding analytic solutions for time-to-go is generally intractable except for rare cases. This means that solving the impact-time-control problem requires computationally-expensive numerical integration to find $t_{go}$. To this end, most existing guidance laws utilized some assumptions to approximate the time-to-go and hence the performance will degrade in realistic scenarios. From extensive numerical simulations, we also demonstrate that conventional ITCG algorithms with approximate analytic time-to-go cannot guide the missile to intercept the target in some scenarios.
\end{remark}

\section{Computational Impact-Time-Control Guidance Algorithm}
\label{sec:2}

To solve the impact time control problem, we propose a learning-based prediction-correction framework to design a computational guidance algorithm, which is formulated as a composite command, i.e.,
\begin{equation}
	a_M=a_0+a_b
\end{equation}
where the baseline command $a_0$ is utilized to provide zero miss distance for target interception and the biased command $a_b$ is developed to nullify the impact time error.

Without loss of generality, we choose the gravity-compensated energy-optimal PNG as the baseline command, i.e.,
\begin{equation}
	a_0=3v\dot \lambda+g \cos{\theta}
\end{equation}

\begin{figure}[hbt!]
	\centering
	\includegraphics[width=.9\textwidth]{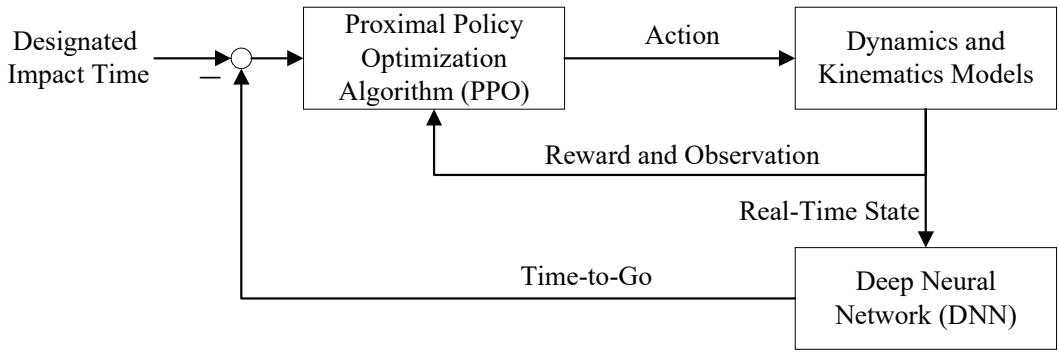}
	\caption{Computational Impact-Time-Control Guidance}
	\label{fig:3}
\end{figure}

The proposed algorithm to design the biased command $a_b$ is composed of two functional blocks: DNN-based real-time predictor and RL-based online corrector. The relationship between these two functional blocks is shown in Fig. \ref{fig:3}.

The predictor leverages DNN to learn the unknown nonlinear mapping from current states to time-to-go under the baseline PNG $a_0$ and can be trained offline. Once the DNN is trained properly, we can then utilize the trained DNN to obtain accurate estimation of time-to-go in real-time, thereby avoiding computationally-expensive numerical integration.

The corrector leverages the state-of-the-art RL algorithm, i.e., Proximal Policy Optimization (PPO), to train a guidance agent that directly outputs the biased command $a_b$ to regulate the impact time error. The rationale behind using RL in the correction step is that the explicit dynamics model of time-to-go is unknown. In RL training, we leverage the trained DNN to predict the time-to-go and use this information as one input to the RL agent. This augmentation transforms the terminal constraint into a state regulation problem and hence resolves the issue of the sparse reward that has been observed in typical RL formulation.

\begin{remark}
Notice that the proposed approach is a general prediction-correction framework for computational guidance law design that can be easily extended to other guidance applications, e.g., impact angle control, terminal velocity control.
\end{remark}

\begin{remark}
Instead of learning from scratch, we utilize a domain-knowledge-aided approach \cite{shin2019computational} in RL training, i.e., we formulate the guidance command as a biased PNG. With this formulation, we only need to train the biased term for impact time error regulation and hence greatly improves the learning effectiveness during the training process.
\end{remark}

\section{The Prediction of Time-to-Go}
\label{sec:3}

This section introduces the details of how to leverage DNN to accurately predict the time-to-go in real time. We first present the architecture of the proposed DNN and then introduce how to collect the training data as well as the training process.

\subsection{The Architecture of DNN}

DNN leverages multiple hidden layers to process features that have been extracted from the input layer and hence provides the possibility of learning complicated nonlinear mappings. For this reason, we utilize the DNN technique to learn the unknown relationship between flight states and time-to-go under PNG. A typical DNN usually consists of input layer, output layer, and multiple fully-connected hidden layers. Each fully-connected layer is composed of a large number of neurons to extract deep features of the input data. Assume that the $k$th layer includes $N$ neurons and the $k+1$th layer consists of $M$ neurons, then the $j$th neuron, $j \in \left\{1,2,\cdots,M \right\}$, in the $k+1$th layer processes the data as \cite{lecun2015deep}
\begin{gather}
	\label{equation:15}
	y_{j} =f_a\left( z_{j}\right)\\
	\label{equation:16}
	z_{j} =\sum \limits_{i=1 }^{N} {w_{ij} x_{i} +b_{ij}}, \quad i \in \left\{1,2,\cdots,N \right\}
\end{gather}
where $x_i$ represents the output of the $i$th neuron from the $k$th layer; $ f_a $ is the activation function; $ w_{ij} $ and $ b_{ij} $ denote the weights and bias, respectively.

To predict the time-to-go using DNN, it is natural to choose the output of the DNN as $ t_{go} $. According to Eq. \eqref{equation:13}, $ t_{go} $ is a function of moving speed $ v $ and trajectory length $ L $. It is clear that the shape of the interception trajectory is influenced by the heading direction $\theta$ and therefore this information should also be considered in the time-to-go estimation. Notice that the future velocity depends on the aerodynamic characteristics of the airframe. Since the air density changes with the variation of height, the remaining flight time under PNG is also indirectly affected by the interceptor's inertial position $\left( x,y \right)$. In conclusion, the time-to-go $ t_{go} $ can be formulated as a function of moving speed $v$, flight path angle $\theta$ and vehicle's position $\left( x,y \right)$ as
\begin{equation}
	\label{equation:17}
	t_{go}=f_t\left(v,\theta,x,y\right)
\end{equation}

Due to the time-varying and nonlinear properties of the aerodynamic model, finding analytic expression of $f_t$ is intractable. For this reason, we utilize a DNN to learn the unknown mapping from $\left(v,\theta,x,y\right)$ to $t_{go}$. Once the DNN is trained properly, we can then leverage the trained network to predict time-to-go in real time. The DNN that has been used in this paper uses 3 fully-connected hidden layers and each hidden layer is composed of 100 neurons. We leverage the well-known rectified linear units (ReLU) function as the activation function for every neuron due to its fast convergence during the training process \cite{lecun2015deep}. The ReLU function is defined as
\begin{equation}
f_a(x)=\left\{\begin{array}{ll}{x,} & {\text { if } x>0} \\ {0,} & {\text { if } x<0}\end{array}\right.
\end{equation}

\begin{figure}[hbt!]
	\centering
	\includegraphics[width=.7\textwidth]{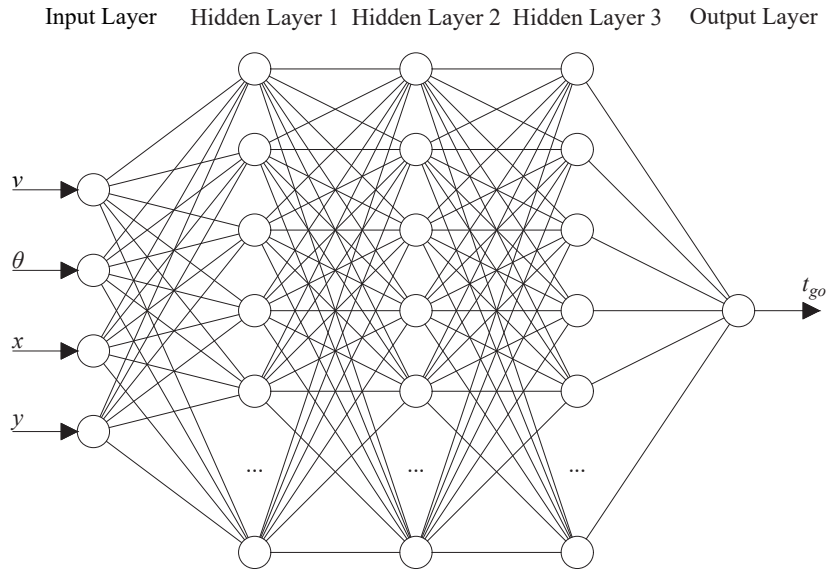}
	\caption{The structure of DNN.}
	\label{fig:4}
\end{figure}

The structure of the constructed DNN is illustrated in Fig. \ref{fig:4}.

\subsection{Training DNN}

The training data in this paper is collected by the simulated flight experiment with realistic aerodynamic models. At every simulation run, we randomly initialize the initial conditions to cover the entire application scenarios. To predict the time-to-go under PNG, we assume that the interceptor is guided by the energy-optimal PNG in each simulation run. Due to the randomness of the initialization, the missile might not be able to intercept the target if the sampled scenario is beyond the physical limits of the interceptor. For this reason, we terminate each simulation run when the relative distance between the missile and the target is less than a threshold or the vertical position of the missile is no longer bigger than zero, i.e., $y \le 0$. Once the simulation is terminated, the missile's states and its flight paths with time information are collected. Then, the time to reach the target from any position in the flight path can be readily obtained by
\begin{equation}
	\label{equation:18}
	t_{go}=t_f-t
\end{equation}

In generating the training samples, we simulate 1000 different interception flight paths and collect 10,156,299 samples with each sample contains the information of network input $\left(v,\theta,x,y\right)$ and output $t_{go}$ pair. These samples are divided into two categories according to the 80/20 principle: the training set is composed of 80\% samples and the other 20\% samples belong to the test set. This principle was demonstrated to outperform other empirical principles in \cite{patgiri2019empirical}. Since the input states have different units and scales, we use their corresponding mean values obtained from all samples to normalize the input states during the training process. Define the DNN is parameterized by $\beta$ and the network parameters are optimized by leveraging the ADAM optimizer \cite{kingma2014adam} with the following loss function
\begin{equation}
	\label{equation:19}
	J\left( \beta \right)=\frac{1}{N_D}\sum\limits_{i=1}^{N_D}\left(\hat{t}_{go,i} -t_{go,i}\right)
\end{equation}
where $ \hat{t}_{go,i} $ is the predicted time-to-go by the DNN from the $i$th sample, and $ t_{go,i} $ is the true time-to-go of the $i$th sample. The symbol $N_D$ denotes the number of samples that have been randomly drawn from the training set  to train the network at each training step.

With objective function \eqref{equation:19}, the network parameter $\beta$ is then updated by the gradient method in a recursive way as
\begin{equation}
\beta_{new}=\beta_{old}+\alpha_{\beta} \nabla_{\beta} J\left( \beta \right)
\end{equation}
where $\alpha_{\beta}$ is the learning rate.

\section{The Correction of Impact Time Error}
\label{sec:4}

Since the dynamics model of the predicted impact time by DNN is unknown, this section introduces the utilization of the state-of-the-art RL algorithm, i.e., PPO, to develop a computational biased command $a_b$ to satisfy the impact time constraint. We first briefly review the PPO algorithm for the completeness of this paper and then presents the RL formulation of the computational guidance agent.

\subsection{Proximal Policy Optimization Algorithm}

The RL problem is often formalized as a Markov Decision Process (MDP) or a partially observable MDP (POMDP). A MDP is described by a five-tuple $\left(\mathcal S, \mathcal O, \mathcal A, \mathcal P, \mathcal R\right)$, where $\mathcal S$ refers to the set of states,  $\mathcal O$ the set of observations, $\mathcal A$ the set of actions, $\mathcal P$ the state transition probability and $\mathcal R$ the reward function. If the process is fully observable, we have $\mathcal S = \mathcal O$. Otherwise, $\mathcal S \ne \mathcal O$. At each time step $t$, an observation $o_t \in \mathcal O$ is generated from the internal state $s_t \in \mathcal S$ given to the agent. The agent utilizes this state to generate an action $a_t \in \mathcal A$ that is sent to the environment based on a specific action policy $\pi\left(a_{t} \mid s_{t}\right)$. The action policy is a function that maps the state to a probability distribution over the actions.  The environment then leverages the action and the current state to generate the next state $s_{t+1}$ with conditional probability $\mathcal P \left( s_{t+1} | s_t,a_t \right)$ and a scalar reward signal $r_t \sim \mathcal R \left( s_t,a_t \right) $. 

The goal of RL is to seek a policy for an agent to interact with an unknown environment while maximizing the expected total reward it received over a sequence of time steps. The total reward in RL is defined as the summation of discounted reward  to facilitate temporal credit assignment as
\begin{equation} \label{eq:2} 
R_t=\sum\limits_{i = t}^N {{\gamma ^{i - t}}{r_i}} 
\end{equation} 
where $\gamma \in \left( 0,1 \right]$ denotes the discount factor.

Given current state $s_t$, the expected total reward is known as the value function
\begin{equation} 
V_{\pi} \left( s_t \right)= \mathbb E _{\pi}\left[ R_t| s_t\right]
\end{equation} 

Many approaches in reinforcement learning also make use of the action-value function 
\begin{equation} 
Q_{\pi} \left( s_t,a_t \right)= \mathbb E _{\pi}\left[ R_t| s_t,a_t\right]
\end{equation}

Generally, RL algorithms can be divided into two categories: the value function method and the policy gradient approach. Value function approaches leverage a nonlinear mapping, e.g., neural network, to approximate the value function and greedily finds the action by iteratively evaluating the value function based on the Bellman optimality condition. These approaches randomly explore the action space and consider all possible actions during each iteration. Therefore, value function algorithms only work with discrete action spaces and the well-known deep Q learning \cite{mnih2015human} belongs to this category. As a comparison, the policy gradient algorithms learn a nonlinear function that directly maps the states to the actions, rather than taking the action that globally maximizes the value function. The action function is updated by following the gradient direction of the value function with respect to the action, thus termed as policy gradient. Thanks to this property, the policy gradient algorithms are applicable to continuous control problems. The PPO algorithm, proposed by OpenAI \cite{schulman2015trust,schulman2017proximal}, is one of the state-of-the-art solutions that belong to the policy gradient approach and hence is suitable to solve the guidance problem.

\begin{figure}[hbt!]
\centering
\includegraphics[width=.7\textwidth]{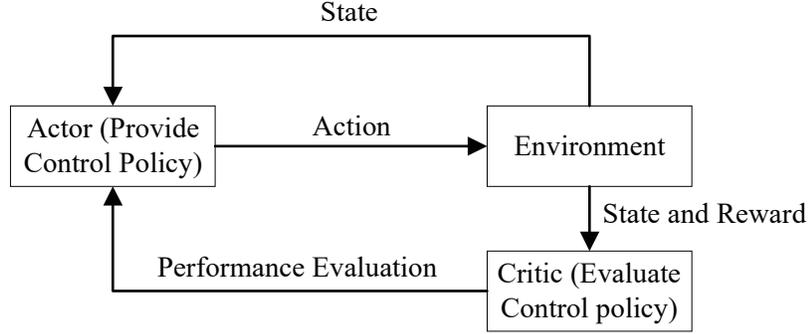}
\caption{The structure of typical PPO.}
\label{fig:6}
\end{figure}

PPO utilizes a typical actor-critic structure, as is shown in Fig. \ref{fig:6}, where the actor generates the control action based on the current states and the critic approximates the value function to evaluate the performance of the action. Both the actor and the critic leverage the gradient method with a batch of $N_s$ samples to update the network parameter. 

\textit{(1) Actor update.} Define the actor network is parameterized by $w$. The original PPO, also termed as Trust Region Policy Optimization (TRPO) \cite{schulman2015trust}, utilizes the following objective function to optimize the actor
\begin{gather}
	\label{equation:23}
	J\left( w \right) = \frac{1}{N_s} \sum_{t=1}^{N_s}\left[\frac{\pi\left(a_{t} \mid s_{t}\right)}{\pi_{old}\left(a_{t} \mid s_{t}\right)} A_{\pi_{\text {old}}}\left(s_{t}, a_{t}\right)\right] \\
	\text {s.t.} \quad \bar{D}_{K L}\left(\pi_{\text {old}}, \pi\right) \leqslant \delta
\end{gather}
where $\pi_{old}\left(a_{t} \mid s_{t}\right)$ denotes the policy that has been optimized in the previous time step; $\bar{D}_{K L}\left(\pi_{\text {old}}, \pi\right)$ is the average Kullback-Leibler divergence (KLD) between $\pi_{\text {old}}$ and $\pi$; $\delta$ is a small constant to constrain the average KLD; $A_{\pi}\left(s_{t}, a_{t}\right)$ is the advantage function, which is defined as the difference between the action-value function and value function, i.e.,
\begin{equation}
A_{\pi}\left(s_{t}, a_{t}\right) = Q_{\pi} \left( s_t,a_t \right) - V_{\pi} \left( s_t\right)
\end{equation}

Notice the average KLD constraint is leveraged to limit the update speed of the policy. This constraint is demonstrated to improve the learning stability during the training process \cite{schulman2015trust}. However, calculating the average KLD is time-consuming and hence the learning speed is constrained by a simple $\texttt{clip}$ function in \cite{schulman2017proximal} as
\begin{equation}
	\label{equation:bb}
	J\left( w \right)=\frac{1}{N_s} \sum_{t=1}^{N_s}\left[\min \left(r_{t}(w) A_{\pi_{o l d}}\left(s_{t}, a_{t}\right), \operatorname{clip}\left(r_{t}(w), 1-\epsilon, 1+\epsilon\right) A_{\pi_{o l d}}\left(s_{t}, a_{t}\right)\right)\right]
\end{equation}
where
 \begin{gather}
        r_t(w) = \frac{\pi\left(a_{t} \mid s_{t}\right)}{\pi_{old}\left(a_{t} \mid s_{t}\right)}\\
        \text{clip}\left[r_t(w),1-\epsilon, 1+\epsilon\right]=\left\{
            \begin{array}{ll}
                1-\epsilon,  & r_t(w)<1-\epsilon \\
                1+\epsilon,  & r_t(w)>1+\epsilon \\
                r_t(w), & 1-\epsilon<r_t(w)<1+\epsilon \\
            \end{array}\right.
\end{gather}
and $\epsilon$ is a small constant to constrain the learning speed.

From Eq. \eqref{equation:bb}, it can be readily observed that the $clip$ function constrains the policy probability ratio with in the range $ (1- \epsilon, 1+ \epsilon) $, thereby indirectly limits the update rate of the policy distribution. With objective function \eqref{equation:bb}, the network parameter $w$ is then updated by moving the policy in the direction of the gradient of $J\left( w \right)$ in a recursive way as
\begin{equation}
w_{new}=w_{old}+\alpha_w \nabla_w J\left( w \right)
\end{equation}
where $\alpha_w$ is the learning rate.

\textit{(2) Critic update.} Define the critic network is parameterized by $\rho$. PPO utilizes the square of the advantage function as objective function in optimizing the critic, i.e.,
\begin{equation}
	\label{equation:bb1}
	J\left( \rho \right)= \frac{1}{N_s} \sum_{t=1}^{N_s}\left\|A_{\pi}\left(s_{t}, a_{t}\right)\right\|^{2}
\end{equation}

With objective function \eqref{equation:bb1}, the network parameter $\rho$ is then updated by the gradient method in a recursive way as
\begin{equation}
\rho_{new}=\rho_{old}+\alpha_{\rho} \nabla_{\rho} J\left( \rho \right)
\end{equation}
where $\alpha_{\rho}$ is the learning rate.

\subsection{RL Formulation of the Impact-Time-Control Guidance Problem}

To develop a computational impact-time-control guidance algorithm by using PPO, we formulate the problem into the RL framework in this subsection. This includes action section, state formulation, reward shaping and network architecture design.

\subsubsection{Action Selection}

As we stated before, the proposed guidance command is based on the general prediction-correction concept. We propose a DNN predictor to obtain accurate time-to-go estimation under PNG and leverage the PPO to develop a RL guidance agent that directly generates the biased command $a_b$ to nullify the impact time error. For this reason, the agent action is naturally defined as
\begin{equation}
	\label{equation:27}
	\text{action}=a_b, \quad |a_b| \le a_{\max}
\end{equation}

Notice that the magnitude of the biased command should be constrained, ensuring that the PNG command plays the dominant role to guarantee target capture. Since the impact-time-control guidance algorithm is normally developed for anti-ship or other air/surface-to-surface missiles, the maximum permissible value of the biased term is limited to $3g$, i.e., $a_{\max}=3g$, to cater for physical constraint.

\subsubsection{State Formulation}

The environmental states are utilized as the inputs to both the actor network and the critic network of PPO. The first consideration in state formulation for our problem is the impact time error since the main objective of the biased command $a_b$ is to nullify the impact time error to satisfy the time-of-arrival constraint. Let $\varepsilon _{t}$ be the impact time error as
\begin{equation}
	\label{equation:28}
	\varepsilon _{t}=t_{d}-\left(t+{\hat t}_{go}\right)
\end{equation}
where ${\hat t}_{go}$ denotes that predicted time-to-go by the proposed DNN.

Notice that the inertial position of the interceptor affects the Mach number and air density. This will, in turn, influence the interception time since the moving speed of missile depends on the aerodynamic characteristics of the airframe. Considering the fact the heading direction of the vehicle determines the trajectory shape and hence indirectly affects the impact time, the state of the $t$th training sample is defined as
\begin{equation}
	\label{equation:29}
	s_{t}=\left(v_t, \theta_{t}, x_{t}, y_{t},\varepsilon_{t}\right)
\end{equation}

As different states have different scales and units, we normalize the engagement states to ensure unit-less states that belong to approximately the same scale. This normalization procedure is shown to be of paramount importance for our problem and helps to increase the training efficiency. Since the training DNN collects the samples of $ (v_t, \theta_{t}, x_{t}, y_{t}) $, we use their corresponding mean values to normalize these states. Due to the lack of samples of $\varepsilon_{t}$, the impact time error is normalized by a tuning parameter $\hat{\varepsilon}$. With the normalization, the environmental states are reformulated as
\begin{equation}
	\label{equation:30}
	state=\left(\frac{v_t}{\bar{v}}, \frac{\theta_t}{\bar{\theta}}, \frac{x_t}{\bar{x}}, \frac{y_t}{\bar{y}},\frac{\varepsilon_t}{\hat{\varepsilon}}\right)
\end{equation}
where the variable with $\bar \cdot$ denotes its mean value and $\hat{\varepsilon}$ stands for the normalization parameter of the impact time error $\varepsilon_{t}$.

\subsubsection{Reward Shaping}

The most important and challenging part in the RL formulation is the design of a proper reward function because this function determines the learning efficiency and guarantees the convergence of the training process. To consider $N_o$ different objectives in reward shaping, we can utilize the multi-objective optimization method to formulate the reward function, i.e.,
\begin{gather}
	\label{equation:31}
	r_t=\sum_{i=1}^{N_o}{a_i r_i}
\end{gather}
where $r_i$ denotes the $i$th reward; $a_i$ is the weight coefficient of the $i$th reward and satisfies the following condition
\begin{gather}
	\sum_{i=1}^{N_o}{a_i}=1
\end{gather}

In reward shaping, we consider the following three objectives to cater for impact time constraint and ensure target interception.

\textit{(1) Impact time error.} The essence of impact-time-control guidance is a finite-time tracking control problem, in which the impact time error is the tracking error. One recent work \cite{he2018optimality} discussed the optimal convergence pattern of the tracking error for general guidance law design. The optimal error dynamics is defined as
\begin{equation}
	\label{equation:33}
	\dot{\varepsilon_t}+\dfrac{k}{t_{go}} \varepsilon_t=0,\quad k>0
\end{equation}
which gives the following analytic solution
\begin{equation}
	\label{equation:34}
	\varepsilon_t=\varepsilon_{t,0} \left( \dfrac{t_{go}}{t_f } \right)^k
\end{equation}
where $\varepsilon_{t,0}$ denotes the initial value of the tracking error $\varepsilon_t$.

From Eq. \eqref{equation:34}, it is clear that the tracking error $\varepsilon_t$ converges to zero at the time of impact. According to \cite{he2018optimality}, error dynamics \eqref{equation:33} minimizes a meaningful performance index, which helps to tune the guidance gain $k$. To mimic this optimal error dynamics, we consider the following reward for the impact time error
\begin{equation}
	\label{equation:36}
	r_1=e^{-\varepsilon_r^2}
\end{equation}
where the auxiliary variable $\varepsilon_r$ is defined as
\begin{equation}
	\label{equation:35}
	\varepsilon_r=\dfrac{\varepsilon_t}{t_{go}}
\end{equation}

\textit{(2) Relative distance.} Notice that there are infinite paths that can extend the flight time to satisfy the impact time constraint. However, some of them cannot satisfy the interception constraint since increasing the flight time will temporarily enforce a detour maneuver away from the target. For this reason, the second reward penalizes the relative distance between the missile and the target to encourage the missile to intercept the target, i.e.,
\begin{equation}
	\label{equation:37}
	r_2=e^{-\frac{R}{\bar{R}}}
\end{equation}
where $ \bar{R} $ represents the normalized coefficient to ensure that the first two rewards belong to similar scale.

\begin{figure}[hbt!]
	\centering
	\includegraphics[width=.6\textwidth]{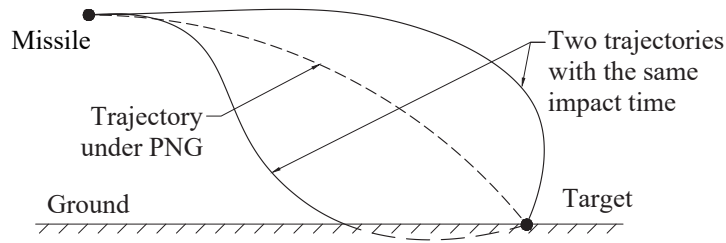}
	\caption{Illustration of different trajectories with the same impact time.}
	\label{fig:illustration}
\end{figure}

\textit{(3) Altitude.} Notice that there are different trajectories that can cater for the constraint of target interception with desired impact time. Hence, the missile might approach the ground before intercepting the target, see Fig. \ref{fig:illustration} for an illustration example, where the dashed line stands for the interception trajectory with PNG and the two solid lines represents the interception trajectories that have the same impact time. From this figure, we can note that if the missile approaches the ground before intercepting the target, the mission will fail and hence we penalize the flight altitude in the third reward as
\begin{equation}
	\label{equation:38}
	r_3=e^{\frac{-\left(y-R\right)^2}{\sigma^2}}
\end{equation}
where $\sigma$ denotes the normalization constant of $\left(y-R\right)$.

In summary, the reward function is defined by combining Eqs. \eqref{equation:36}, \eqref{equation:37} and \eqref{equation:38} as the following weighted sum
\begin{equation}
	\label{equation:39}
	r_t=a_{1} e^{-\varepsilon_{r}^{2}}+a_{2} e^{-\frac{R}{\bar{R}}}+a_{3} e^{-\frac{(y-R)^{2}}{\sigma^{2}}}
\end{equation}

\subsubsection{Network Architecture}

Inspired by the original PPO algorithm \cite{schulman2017proximal}, both the actor and the critic are represented by four-layer fully-connected neural networks. Note that this four-layer network architecture is commonly utilized in deep reinforcement learning applications \cite{henderson2018deep}. The layer sizes of these two networks are summarized in Table \ref{tab:nn}. Except for the actor output layer, each neuron in other layers is activated by a ReLU function, which provides faster processing speed than other nonlinear activation functions due to the linear relationship property. The output layer of the actor network is activated by the $\tanh$ function, which is given by
\begin{equation} 
g\left(x\right)=\frac{e^{x}-e^{-x}}{e^{x}+e^{-x}}
\end{equation}

\begin{table}[h]
    \caption{\label{tab:nn} Network layer size.}
	\centering
    \begin{tabular}{c|c|c}
    \hline
    Layer          & Actor network                                & Critic network                \\ \hline
    Input layer    & 5 (Size of states)                            & 5 (Size of states)             \\
    Hidden layer 1 & 64                                           & 64                            \\
    Hidden layer 2 & 64                                           & 64                            \\
    Output layer   & 1 (Size of action) & 1 (Scalar value function) \\ \hline
    \end{tabular}
\end{table}

The benefit of the utilization of $\tanh$ activation function in actor network is that it can prevent the control input from saturation as the actor output is constrained by $ \left[-1, 1\right] $. The output layer of the actor network is scaled by a constant $a_{\max}$ to constrain the biased command within the range $ \left[-a_{\max}, a_{\max}\right] $.

\subsubsection{Network Training}

In the training process, we use a buffer to store a batch of $N_s$ transition experience samples. A transition experience $ e_t $ is defined as
\begin{equation}
	\label{equation:40}
	e_t=(s_t, a_t, r_t, s_{t+1})
\end{equation}

We assume that the action policy is subject to a Gaussian distribution, i.e.,
\begin{equation} 
a_t \sim {\mathcal N}\left(\mu_a,\sigma_a\right)
\end{equation}
where ${\mathcal N}\left(\mu_a,\sigma_a\right)$ denotes a Gaussian distribution with its mean and standard deviation as $\mu_a$ and $\sigma_a$, respectively.

For simplicity, the standard deviation $\sigma_a$ is assumed as constant and this parameter is trained with the actor parameter $w$ in an integrated manner by the gradient method. Hence, the output of the actor network is the mean of action, i.e., $\mu_a$. Both the critic and the actor networks are optimized by the ADAM algorithm with an episodic manner. At the beginning of each episode, the states of the environment are initialized randomly and the experience buff is initialized as a zero set. Once $N_s$ samples are stored in the experience buffer, the network parameters are then updated by using these $N_s$ samples and we empty the experience buffer. If the number of time steps reaches the maximum value $T_{\max}$ or the missile approaches the ground/target, the episode is terminated.

\section{Simulation Results}
\label{sec:5}

In this section, the performance of the proposed computational ITCG algorithm is evaluated by numerical simulations. We first investigate the accuracy of the DNN time-to-go estimator and then analyze the performance of the PPO-based impact time error corrector using Monte-Carlo simulations.

\subsection{The Performance of DNN Predictor}

The training and test data is collected by using different interception trajectories under energy-optimal PNG. The initial conditions of these simulated scenarios are summarized in Table \ref{tab:table1}. Notice that the missile's initial position, initial flight path angle and initial moving speed are randomly sampled from a uniform distribution within the ranges that have been specified in Table \ref{tab:table1}. The aerodynamic characteristics of the considered airframe with respect to different Mach numbers are presented in Table \ref{tab:table2} and the aerodynamic coefficients of other operational points are obtained by interpolation. The reference area of the interceptor is $ S=0.0572556m^2 $ and the gravitational acceleration is constant as $ g=9.81m/s^2 $. To meet the physical constraints of the missile, the AoA command is limited by a maximum value of $15^{\circ}$. As stated before, we utilize the mean of all samples to normalize the data collected from the simulated scenarios. The hyper parameters in training the DNN are summarized in Table \ref{tab:dnn}.

\begin{table}[hbt!]
	\caption{\label{tab:table1} Initial conditions.}
	\centering
	\begin{tabular}{clccc}
		\hline
		Parameter & Description                                      & Value or Interval & Units \\ \hline
		$ x_0 $    & Initial missile position in the $x$ direction & (-30,-10)         & km    \\
		$ y_0 $    & Initial missile position in the $y$ direction & (10,30)           & km    \\
		$ x_T $    & Target position in the $x$ direction   & 0                 & km    \\
		$ y_T $    & Target position in the $y$ direction   & 0                 & km    \\
		$ v_0 $    & Initial missile velocity                     & (200,300)         & m/s   \\
		$ \theta_0 $ & Initial flight path angle                        & (0,45)            & deg   \\
		$ m $    & The mass of the missile                                     & 200               & kg    \\ \hline
	\end{tabular}
\end{table}

\begin{table}[hbt!]
	\caption{\label{tab:table2} Aerodynamic coefficients and derivatives.}
	\centering
	\begin{tabular}{cccc}
		\hline
		Mach Number  & $C_L^{\alpha}/rad^{-1}$ & $C_{D0}$ & $C_D^{\alpha^2}/rad^{-2}$ \\ \hline
		0.4 & 39.056                  & 0.4604   & 39.072                    \\
		0.6 & 40.801                  & 0.4682   & 39.735                    \\
		0.8 & 41.372                  & 0.4635   & 39.242                    \\
		0.9 & 42.468                  & 0.4776   & 40.531                    \\ \hline
	\end{tabular}
\end{table}

\begin{table}[hbt!]
	\caption{\label{tab:dnn} Hyper parameter settings in training DNN.}
	\centering
	\begin{tabular}{clcc}
	 \hline
        Parameter      & Description                   & Value  \\ \hline
	 $T_t$      & Number of training steps	 & 100,000    \\
	 $N_D$         & Number of samples for every training step          & 1000 \\
	 $a_\beta$           & Learning rate  	    & 0.001 \\ \hline
	\end{tabular}
\end{table}

\begin{figure}[hbt!]
	\centering
	\includegraphics[width=0.8\textwidth]{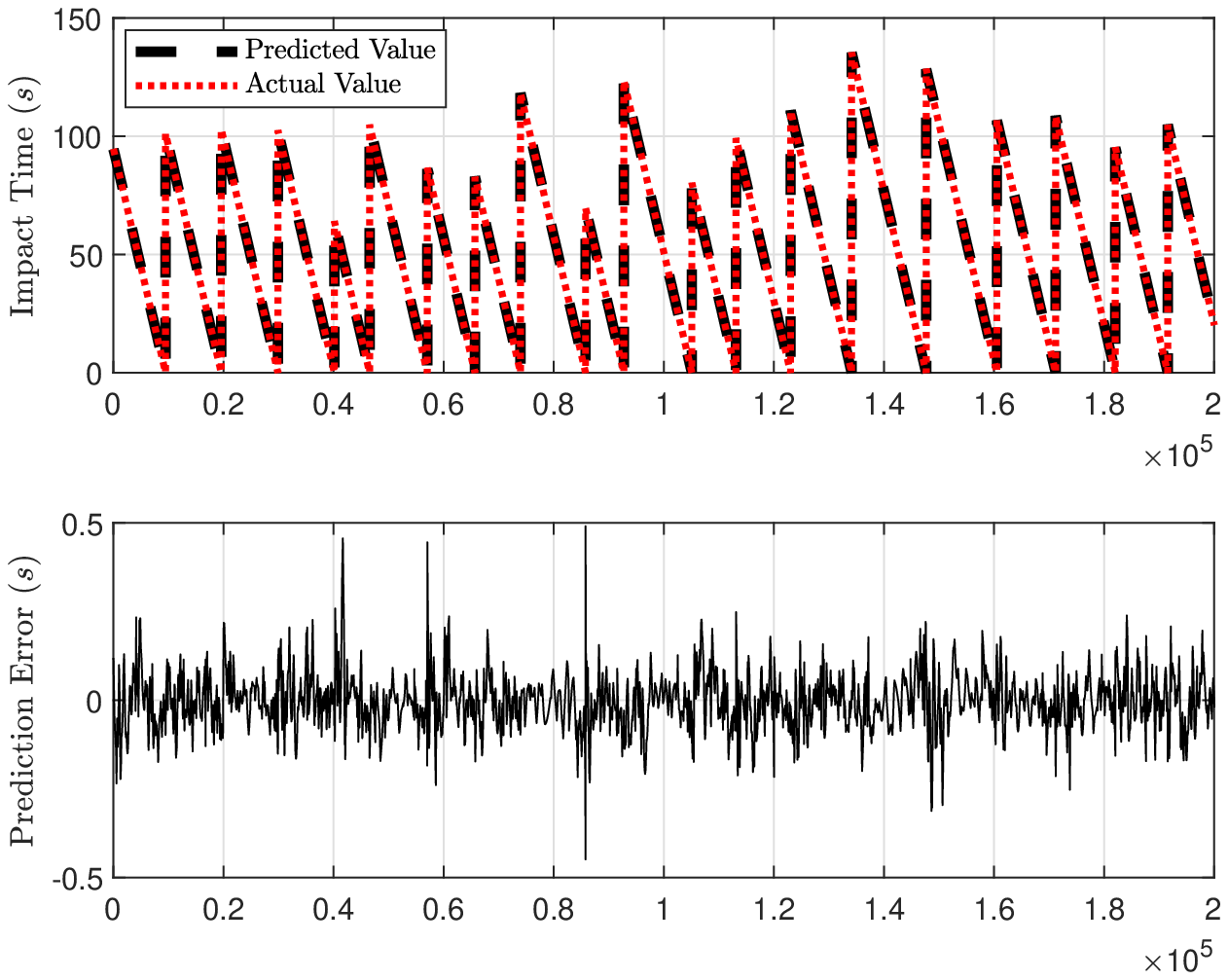}
	\caption{Test results of the proposed DNN predictor.}
	\label{fig:8}
\end{figure}

Table \ref{fig:8} presents the test results of 20000 test samples generated by the proposed DNN predictor. From this figure, we can readily observe that the proposed DNN provides accurate impact time prediction and the maximum prediction error is smaller than $0.5s$. The statistical characteristics of the DNN prediction error obtained from all test samples are summarized in Table \ref{tab:table3}, where the coefficient of determination (CR) is determined by
\begin{equation}
	\text{CR}=\dfrac{\left(N_t \sum\limits_{i=1}^{N_t } \hat{t}_{go,i} t_{go,i}-\sum\limits_{i=1}^{N_t } \hat{t}_{go,i}\sum\limits_{i=1}^{N_t } t_{go,i}\right)^{2}}{\left[N_t  \sum\limits_{i=1}^{N_t } \hat{t}_{go,i}^{2}-\left(\sum\limits_{i=1}^{N_t } \hat{t}_{go,i}\right)^{2}\right]\left[N_t  \sum\limits_{i=1}^{N_t } t_{go,i}^{2}-\left(\sum\limits_{i=1}^{N_t } t_{go,i}\right)^{2}\right]}
\end{equation}
where $N_t$ denotes the number of test samples.

The results in Table \ref{tab:table3} demonstrate the effectiveness of the proposed DNN predictor: accurate prediction with small error variation. Notice that the CR result indicates the proposed DNN is reliable and hence can be utilized in real-time prediction of time-to-go.
\begin{table}[hbt!]
	\caption{\label{tab:table3} The statistical characteristics of the DNN prediction error.}
	\centering
	\begin{tabular}{cccc}
		\hline
		Mean Value & Standard Deviation & Maximum Value & Coefficient of Determination \\ \hline
		-0.0048 & 0.0744 & 0.4835 & 1.0000  \\ \hline
	\end{tabular}
\end{table}

\subsection{The Performance of PPO Corrector}

\subsubsection{Training the PPO Corrector}

The initial conditions of the scenarios that have been utilized in PPO training are the same as described in Table \ref{tab:table1}. To make the impact-time-control problem feasible, the desired impact time is randomly set as 1.1 to 1.2 times of the predicted result for every episode. The hyper parameters that are utilized in PPO training for our problem are summarized in Table \ref{tab:table4}. Notice that the tuning of hyper parameters imposes great effects on the performance of PPO and this tuning process is not consistent across different ranges of applications, i.e., different works utilized different set of hyper parameters for their own problems. For this reason, we tune these hyper parameters for our guidance problem based on several trial and error tests.

\begin{table}[hbt!]
	\caption{\label{tab:table4} Hyper parameter settings in training PPO.}
	\centering
	\begin{tabular}{clcc}
	 \hline
        Parameter      & Description                   & Value  \\ \hline
	 $\epsilon$      & Ratio clipping	            & 0.2    \\
	 $a_w$         & Actor learning rate           & 0.0001 \\
	 $a_\rho$           & Critic learning rate  	    & 0.0002 \\
        $\gamma$        & Discounting factor  	        & 0.995  \\
        $N_s$             & Size of the experience buffer    & 256    \\
	 $T_{\max}$       & Maximum permissible time steps of  one episode	 	& 400    \\
        $E_{\max}$       & Maximum permissible episodes  & 500 	 \\
        $a_{1}$         & The weight of the 1st reward & 0.9    \\
        $a_{2}$         & The weight of the 2nd reward & 0.09   \\
        $a_{3}$         & The weight of the 3rd reward & 0.01   \\
        $\hat{\varepsilon}$ & The normalization parameter for $ \varepsilon_{t} $ & 2 \\
        $\bar{R}$ & The normalization parameter for the 2nd reward & $1.6\times10^4$ \\
        $\sigma$ & The normalization parameter for the 3rd reward & $115$\\ \hline
	\end{tabular}
\end{table}

\begin{figure}[hbt!]
	\centering
	\includegraphics[width=0.5\textwidth]{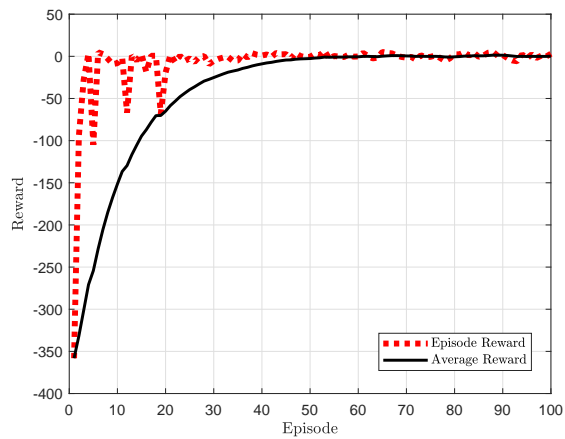}
	\caption{Learning curve of the proposed PPO.}
	\label{fig:9}
\end{figure}

\begin{figure}[hbt!]
	\centering
	\includegraphics[width=0.5\textwidth]{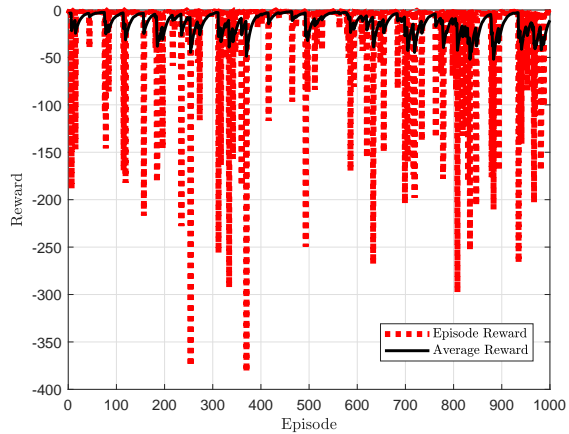}
	\caption{Learning curve without using DNN.}
	\label{fig:ppo2}
\end{figure}

\begin{figure}[hbt!]
	\centering
	\includegraphics[width=0.5\textwidth]{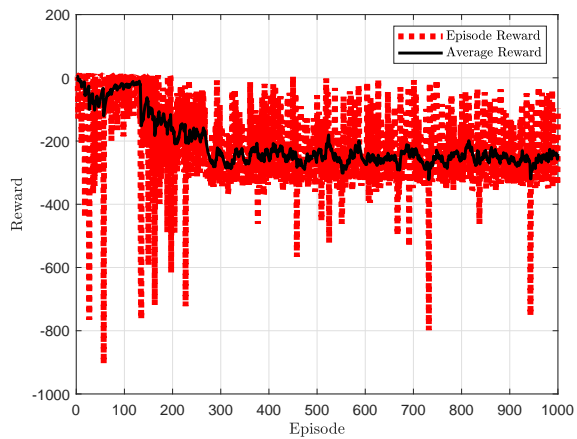}
	\caption{Learning curve of PPO with approximate PNG time-to-go.}
	\label{fig:ppo3}
\end{figure}

The convergence pattern of the reward function in training the PPO is shown in Fig. \ref{fig:9}. From this figure, it can be clearly observed that the average reward of the proposed computational ITCG algorithm converges to the steady-state within 100 episodes. To show the effectiveness of the proposed DDN+PPO concept, we also conduct comparison simulations with training PPO from scratch and training PPO with approximate PNG time-to-go. Training PPO from scratch refers to the concept without using DNN to predict the time-to-go and we give a positive reward once the missile intercept the target with desired time. The learning curve of this concept is plotted in Fig. \ref{fig:ppo2}, which demonstrates that the learning process is not stable due to the effect of sparse reward. This can be attributed the fact that the probability of intercepting the target with a desired impact time under random initial conditions is very low. Figure \ref{fig:ppo3} presents the learning curve of training PPO with approximate PNG time-to-go, which is determined by \cite{jeon2006impact}
\begin{equation}
\label{equation:tgo}
        \hat{t}_{go}=\frac{\left[1+\frac{\left(\theta-\lambda\right)^2}{10}\right]R}{v}
\end{equation}

The results in Fig. \ref{fig:ppo3} show that the average reward cannot be maximized if we use approximate time-to-go estimations for realistic scenarios. The reason is that time-to-go estimation \eqref{equation:tgo} assumes that the moving speed of the interceptor is constant and ignores the effect of gravity. This demonstrates the importance of augmenting the DNN predictor into the PPO training process to ensure stable learning.

\subsubsection{Performance Analysis of the Proposed Computational ITCG Algorithm}

We consider a fixed initial condition to investigate the performance of the proposed computational ITCG algorithm. The initial condition is set as
\begin{equation}
	\label{equation:47}
	x_0=-20km, \quad y_0=20km, \quad v_0=200m/s, \quad \theta_0=0^{\circ}
\end{equation}

\begin{figure}[hbt!]
	\centering
	\includegraphics[width=1\textwidth]{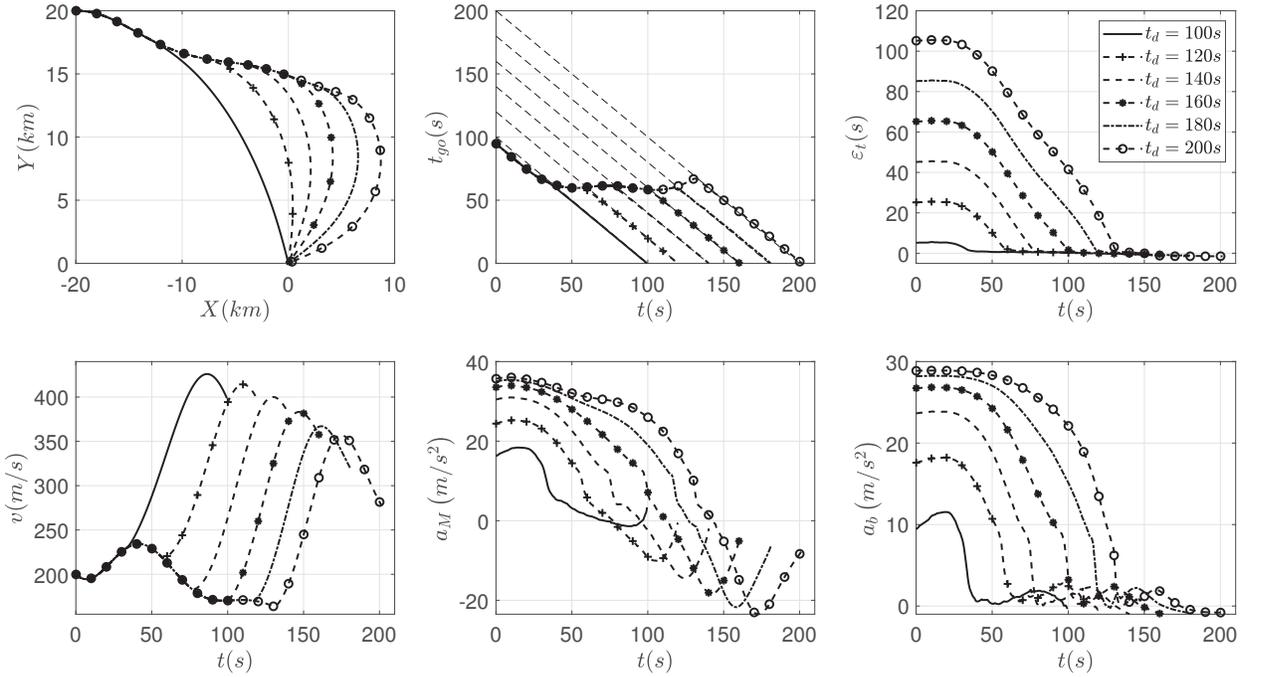}
	\caption{Performance of the proposed ITCG algorithm with different desired impact time.}
	\label{fig:10}
\end{figure}

The simulation results, including interception trajectory, time-to-go history, convergence of impact time error, moving speed, guidance command and biased command, under the proposed computational ITCG algorithm with different desired impact time $t_d=100s,120s,140s,160s,180s,200s$ are presented in Fig. \ref{fig:10}. From this figure, it can be clearly observed that the missile can successfully intercept the target at the desired time under the proposed computational ITCG algorithm. The interception trajectory becomes more curved with the increase of the desired impact time and hence requires large biased acceleration command to nullify the impact time error. The results also indicate that the guidance command of the proposed algorithm converges to around zero at the time of impact, thereby providing enough operational margins to cope with undesired external disturbances. The velocity profile shown in Fig. \ref{fig:10} demonstrates that the moving speed significantly changes with different interception trajectories. This indirectly means that the ITCG algorithms under constant-speed assumption cannot cater for practical scenarios. To see this, we further conduct numerical comparisons with PNG and existing analytic ITCG algorithms \cite{jeon2006impact,tahk2018impact} with desired impact time set as $t_d=120s$. The guidance commands of these two analytic guidance laws are formulated as
\begin{gather}
	\text{ITCG1 in \cite{jeon2006impact}}: a_M = 3v\dot{\lambda}+\dfrac{-120v^5}{3v\dot{\lambda}R^3}\left(t_d-t-\hat{t}_{go}\right)+g \cos{\theta}  \\
	\text{ITCG2 in \cite{tahk2018impact}}: a_M = -\dfrac{3v^2}{R}\left(\theta-\lambda\right)+\dfrac{100v^2}{R\left(\theta-\lambda\right)}\dfrac{t_d-t-\hat{t}_{go}}{t_d-t}+g \cos{\theta}
\end{gather}

\begin{figure}[hbt!]
	\centering
	\includegraphics[width=1\textwidth]{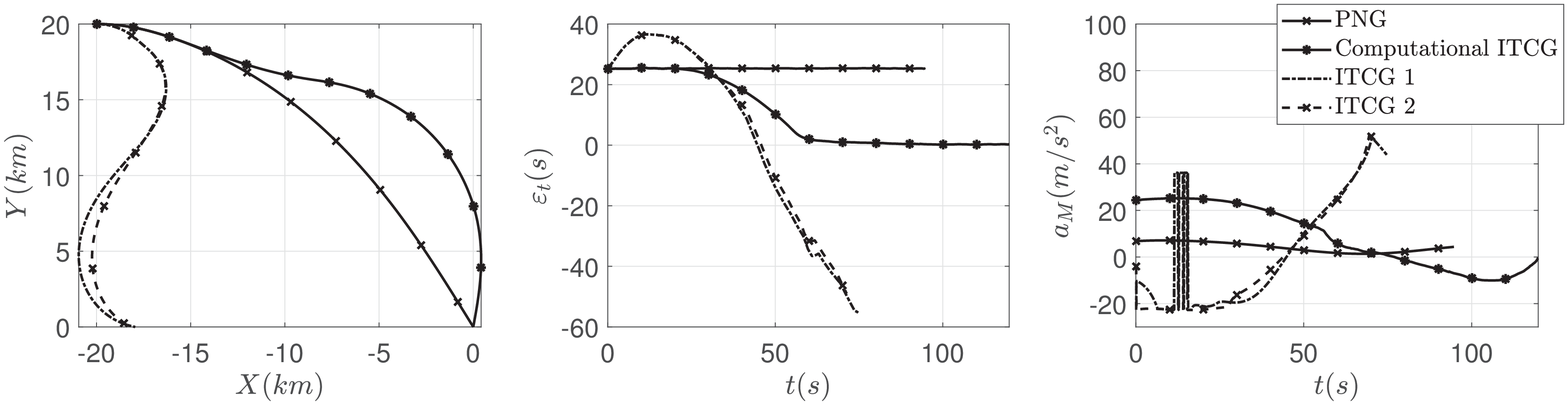}
	\caption{Comparison results with existing analytic ITCG algorithms.}
	\label{fig:com}
\end{figure}

The simulation results, obtained from different guidance laws, are presented in Fig. \ref{fig:com}, which demonstrate that both PNG and the proposed computational ITCG algorithm can successfully guide the missile to intercept the target. Due to the introduced biased term, the proposed computational ITCG algorithm can be leveraged to satisfy the impact time constraint and the recorded interception time is $ t_f=119.8s $. This requires additional control effort to make detour maneuvers to increase the flight time compared to PNG, as confirmed by the profile of the acceleration command. As a comparison, both analytic ITCG laws \cite{jeon2006impact,tahk2018impact} fail to intercept the target since these two algorithms are derived under ideal conditions.

\subsubsection{Monte-Carlo Analysis of the Proposed Computational ITCG Algorithm}

\begin{figure}[hbt!]
	\centering
	\includegraphics[width=1\textwidth]{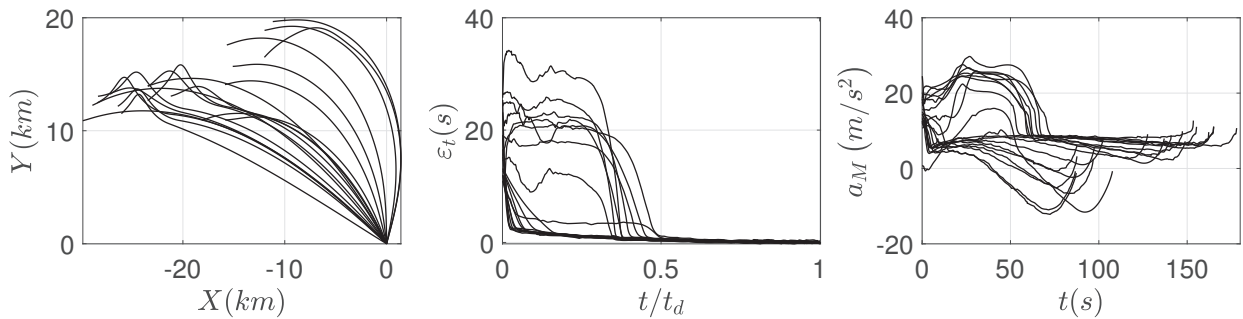}
	\caption{Monte-Carlo simulation results with random initial conditions.}
	\label{fig:13a}
\end{figure}

To test the proposed computational ITCG algorithm under various conditions, Monte-Carlo simulations are performed with random initial conditions and random desired impact time. The Monte-Carlo simulation results, including interception trajectory, impact time error and acceleration command, are shown in Fig. \ref{fig:13a}, where the time duration in plotting the impact time error is normalized by $t_d$ to better show the convergence patter. It follows from Fig. \ref{fig:13a} that the impact time error under the proposed computational ITCG algorithm converges to zero at the time of impact with random scenarios. This clearly conforms with the physical insights of the reward function that we have developed. The statistical characteristics of the impact time error for 100 Monte-Carlo are summarized in Table \ref{tab:table5}. The results reveal that the proposed approach provides accurate impact time control with satisfactory performance, i.e., the impact time error is smaller than $0.5s$ for more than 85\% scenarios.

\begin{table}[hbt!]
	\caption{\label{tab:table5} The statistical characteristics of the impact time error.}
	\centering
	\begin{tabular}{ccc}
		\hline
		Mean Value   & Standard Deviation   & Maximum Value \\ \hline
		-0.2145 & 0.9264 & 3.9100   \\ \hline
	\end{tabular}
\end{table}

\begin{figure}[hbt!]
	\centering
	\includegraphics[width=0.5\textwidth]{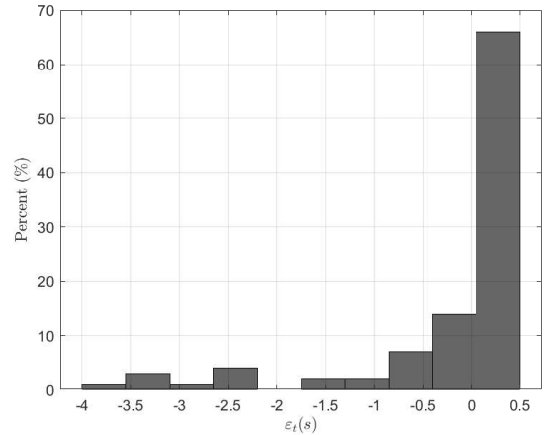}
	\caption{The distribution of the impact time error.}
	\label{fig:13}
\end{figure}

\section{Conclusion}

This paper proposes a computational impact-time-control guidance algorithm based on a general prediction-correction concept. The deep neural network is leveraged as a real-time predictor to estimate the time-to-go under PNG with realistic aerodynamic models. The biased command to nullify the impact time error is developed by utilizing the emerging reinforcement learning techniques. Extensive numerical simulations reveal that the proposed approach provides promising performance in implementation under realistic scenarios.


\bibliographystyle{IEEEtran}
\bibliography{ITCG}

\end{document}